\documentclass[preprint]{elsarticle}
\usepackage{xcolor}
\usepackage{multirow}
\usepackage{url}
\usepackage{amsmath}
\usepackage{amsfonts}
\usepackage{amssymb}

\definecolor{green}{RGB}{0,0,0}
\newcommand{\green}[1]{{\color{green}#1}}
\usepackage[linesnumbered,ruled,vlined]{algorithm2e}

\AtBeginDocument{%
  \providecommand\BibTeX{{%
    \normalfont B\kern-0.5em{\scshape i\kern-0.25em b}\kern-0.8em\TeX}}}

\begin{document}

\title{\textsc{roadscene2vec}: A Tool for Extracting and Embedding Road Scene-Graphs}

\author{Arnav Vaibhav Malawade\fnref{fn1}\corref{cor1}}
\ead{malawada@uci.edu}

\author{Shih-Yuan Yu\fnref{fn1}\corref{}}
\ead{shihyuay@uci.edu}
\author{Brandon Hsu\corref{}}
\ead{bdhsu@uci.edu}
\author{Harsimrat Kaeley\corref{}}
\ead{kaeleyh@uci.edu}
\author{Anurag Karra\corref{}}
\ead{karraa@uci.edu}
\author{Mohammad Abdullah Al Faruque\corref{}}
\ead{alfaruqu@uci.edu}
\address{Department of Electrical Engineering \& Computer Science, University of California - Irvine, Irvine, CA 92697, USA}
\fntext[fn1]{Both authors contributed equally to this research.}
\cortext[cor1]{Corresponding Author}

\begin{abstract}
Recently, road scene-graph representations used in conjunction with graph learning techniques have been shown to outperform state-of-the-art deep learning techniques in tasks including action classification, risk assessment, and collision prediction.
To enable the exploration of applications of road scene-graph representations, we introduce \textsc{roadscene2vec}: an open-source tool for extracting and embedding road scene-graphs. 
The goal of \textsc{roadscene2vec} is to enable research into the applications and capabilities of road scene-graphs by providing tools for generating scene-graphs, graph learning models to create spatio-temporal scene-graph embeddings, and tools for visualizing and analyzing scene-graph-based methodologies. 
The capabilities of \textsc{roadscene2vec} include (i) customized scene-graph generation from either video clips or data from the CARLA simulator, (ii) multiple configurable spatio-temporal graph embedding models and baseline CNN-based models, (iii) built-in functionality for using graph and sequence embeddings for risk assessment and collision prediction applications, (iv) tools for evaluating transfer learning, and (v) utilities for visualizing scene-graphs and analyzing the explainability of graph learning models. We demonstrate the utility of \textsc{roadscene2vec} for these use cases with experimental results and qualitative evaluations for both graph learning models and CNN-based models. \textsc{roadscene2vec} is available at \url{https://github.com/AICPS/roadscene2vec}.
\end{abstract}

\maketitle


\section{Introduction}
Autonomous Vehicles (AVs) are expected to revolutionize personal mobility, logistics, and road safety~\cite{litman2017autonomous}.
However, recent accidents involving Tesla Autopilot and Uber's self-driving cars indicate that the development of safe and robust AVs remains a difficult challenge~\cite{NTSB2019uber, NTSB2018, NTSB2019}.
Current statistics indicate that perception and prediction errors were factors in over 40\% of driver-related crashes between conventional vehicles \cite{mueller2020humanlike}, leading both researchers and industry leaders to race to address these problems via advanced AV perception systems.
Until recently, most AV perception architectures relied entirely on either (i) deep learning techniques using Convolutional Neural Networks (CNNs) and Multi-Layer Perceptrons (MLPs) \cite{yurtsever2019risky, bojarski2016end, tao2021stereo, xiao2020attention}; or (ii) model-based methods, which use known road geometry information and vehicle trajectory models to estimate the state of the road scene \cite{sontges2018worst, nister2019safety}.
\green{Although these approaches have been successful in typical use cases, they cannot obtain a high-level, human-like understanding of complex road scenarios. This limitation is due to their inability to explicitly capture inter-object relationships or the overall structure of the road scene.}

\begin{figure}[!t]
    \includegraphics[clip, trim=0 180 0 130, width=1.0\linewidth]{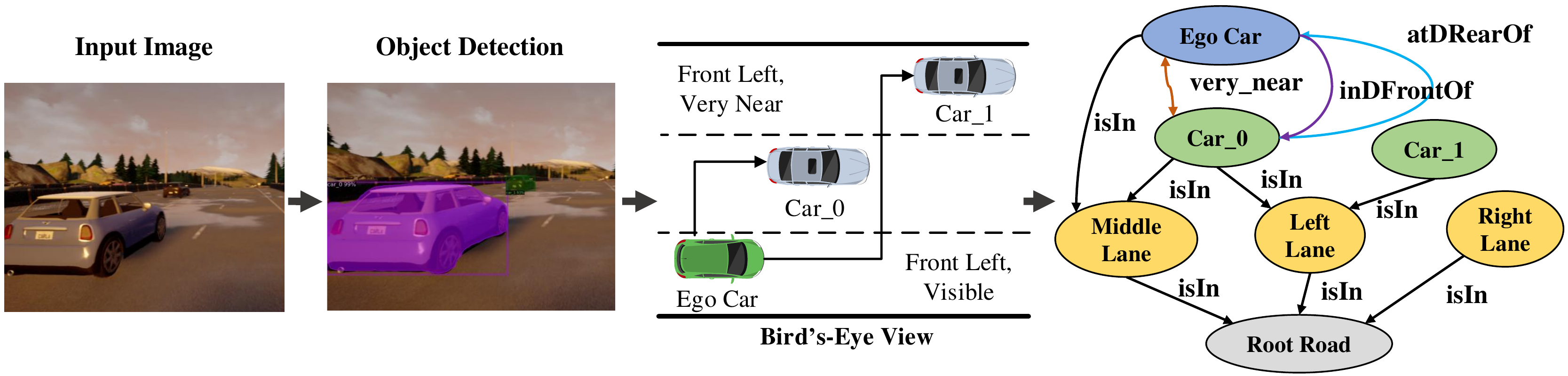}
    \caption{How camera data can be used to construct a road \textit{scene-graph} representation.}
    \label{fig:scene_graph}
\end{figure}

Research has suggested that humans rely on cognitive mechanisms to identify the structure of a scene and reason about inter-object relations when performing complex tasks (e.g., identifying risk) \cite{battaglia2018relational}. 
As such, \textcolor{green}{the capability to capture and identify the complex relationships between road objects is \textcolor{green}{critical} in designing an effective human-like AV perception system. }
\textcolor{green}{To address the limitations of existing AV perception methods, several groups have adopted a variant of knowledge graphs known as \textit{scene-graphs} to model the road state and the relationships between objects~\cite{yu2021scene, mylavarapu2020towards, li2019learning}.} 
A \textit{scene-graph} representation encodes rich semantic information of an image or observed scene, essentially bringing an abstraction of objects and their complex relationships as illustrated in Figure~\ref{fig:scene_graph}. 
\textcolor{green}{While each of these related works proposes a different form of \textit{scene-graph} representation, all demonstrate significant performance improvements over conventional perception methods.}
In~\cite{li2019learning}, the authors propose a 3D-aware egocentric spatio-temporal interaction framework that uses both an \textit{Ego-Thing} graph and an \textit{Ego-Stuff} graph, which together encode how the ego vehicle interacts with both moving and stationary objects in a scene, respectively.
In~\cite{mylavarapu2020towards}, the authors propose a pipeline using a multi-relational graph convolutional network (MR-GCN) for classifying the driving behaviors of traffic participants.
\textcolor{green}{The MR-GCN combines spatial and temporal information, including relational information between moving objects and landmark objects.}
\textcolor{green}{Our prior work has demonstrated that a spatio-temporal \textit{scene-graph} embedding can be used to assess the subjective risk of driving maneuvers more effectively than state-of-the-art methods~\cite{yu2021scene}.} 
\textcolor{green}{In addition, our method can better transfer knowledge and is more explainable.}

Although a wide range of \textit{scene-graph} based AV perception approaches have been proposed, each method was developed from scratch, requiring significant time and resource investment by each research group. Although tools exist to perform preprocessing and graph learning (e.g., Pytorch and Pytorch Geometric), to the best of our knowledge, there exists no tool for systematically converting road scenes into \textit{scene-graphs} in this field. As a result, each research group must start developing their \textit{scene-graph} construction methodology from the ground up, wasting time and effort that could be better spent using the resultant \textit{scene-graph} representations to solve more complex research problems. To address this problem, we propose \textsc{roadscene2vec}: a tool for systematically extracting and embedding road \textit{scene-graphs}. \textsc{roadscene2vec} enables researchers to quickly and easily extract scene graphs from camera data, evaluate different graph construction methodologies, \textcolor{green}{and use several different graph learning and machine learning algorithms to generate spatio-temporal graph embeddings for a wide range of AV tasks.}
We envision \textsc{roadscene2vec} to serve the following use cases:
\begin{itemize}
    \item Converting image-based datasets as well as datasets generated by the CARLA simulator \cite{dosovitskiy2017carla} into \textit{scene-graphs}.
    \item Enabling the exploration of different \textit{scene-graph} construction methodologies for a given application via a flexible, reconfigurable, and user-friendly \textit{scene-graph} extraction framework.
    \item Allowing researchers to explore various spatio-temporal graph embedding methods, supporting customized algorithms for further design exploration.
    \item Providing a set of baselines drawn from state-of-the-art works for different AV applications (CNN and CNN-LSTM based algorithms).
    \item We provide \textit{scene-graph} visualization utilities to enhance design space exploration for graph construction. 
\end{itemize}

\textcolor{green}{We target camera data since images are the most rich and detailed modality, providing high-resolution details about the scene as well as color information.} This information can be used for better identifying the context of the scene and relations between participants. If other modalities are added, it is unlikely that much more information will be added to the scene graph; only the robustness of the system and precision of the graph will be improved. Besides, current state-of-the-art AV perception architectures utilizing sensor fusion still have shortcomings \cite{fang2021invisible}. \textcolor{green}{Furthermore, most publicly available AV datasets primarily contain image data.}

\subsection{Novel Contributions}
Our novel contributions for this research community are:
\begin{enumerate}
\item We present \textsc{roadscene2vec}: a flexible \textit{scene-graph} construction and embedding framework that allows researchers to experiment with different graph extraction formulations to find the best one for their problem. 

\item We provide an end-to-end graph learning framework for modeling the \textit{scene-graph} representations. Our framework enables automated experimentation and metrics logging over a wide range of graph learning AV applications. \green{We also provide a graph learning model template defining the core structure and functions used by our framework to facilitate users defining their own models and problems.}

\item We provide many visualization tools and utilities for inspecting and understanding the \textit{scene-graphs} including attention maps, color-coding by classes or relation type, birds-eye view projection, embedding projection, etc. These tools enable users to interpret their results easily without having to design their own visualizer.

\item We provide state-of-the-art CNN-based models drawn from recent AV papers for cross-comparison with graph-learning-based techniques. 

\end{enumerate}

\subsection{Paper Organization}
The rest of our paper is laid out as follows. In Section \ref{sec:related_work} we discuss related works. In Section \ref{sec:roadscene2vec} we introduce the core functionality of our tool and its methodology. In Section \ref{sec:usage} we provide usage examples. In Section \ref{sec:discussion} we demonstrate the practical, real-world value of our tool by evaluating it on several common use cases. Finally, in Section \ref{sec:conclusion} we present our conclusions.
\section{Related Work}
\label{sec:related_work}
In this section, we begin by describing general AV design philosophies. Then we elaborate on graph-based approaches used in scene understanding. Lastly, we briefly discuss existing tools and libraries.

\subsection{AV Design Methodologies}
The two common design approaches for AV systems are (i) end-to-end deep learning architectures~\cite{yurtsever2019survey} and (ii) modular architectures. 
Modular approaches are implemented as a pipeline of separate components for performing each sub-task of the AV (e.g., perception, localization, planning, control). In contrast, end-to-end approaches generate actuator outputs (e.g., steering, brake, accelerator) directly from their sensory inputs~\cite{bojarski2016end}.
One advantage of a modular design approach is the division of a task into an easier-to-solve set of sub-tasks that have been addressed in other fields such as robotics, computer vision, and vehicle dynamics, from which prior knowledge can be leveraged.
However, one disadvantage of such an approach is the complexity of implementing, running and validating the complete pipeline~\cite{yurtsever2019survey}. 
\textcolor{green}{End-to-end approaches can achieve good performance with smaller network size and low implementation costs because they perform feature extraction from sensor inputs implicitly through the network's hidden layers~\cite{bojarski2016end}.}
However, the authors in~\cite{chen2015deepdriving} point out that the needed level of supervision is too weak for the end-to-end model to learn critical control information (e.g., from image to steering angle), so it can fail to handle complicated driving maneuvers or be insufficiently robust to disturbances.

A third approach called the \textit{direct perception} approach was first proposed by DeepDriving~\cite{chen2015deepdriving}. 
In this approach, a set of \textit{affordance indicators}, such as the distance to lane markings and other cars in the current and adjacent lanes, are extracted from an image and serve as an intermediate representation (IR) for generating the final control output.
They show that this IR improves performance for simple driving tasks such as lane following and enables better generalization to real-world environments.
Similarly, \cite{bansal2018chauffeurnet} uses a collection of filtered images as the IR. They state that the IR used in their approach allows the training to be conducted on either real or simulated data, facilitating testing and validation in simulations before testing on a real car.
Moreover, they show that it is easier to synthesize perturbations to the driving trajectory in the IR than at the raw sensor inputs themselves, enabling them to produce non-expert behaviors such as off-road driving and collisions.
The authors in~\cite{yurtsever2019risky} use Mask-RCNN~\cite{he2017mask} to color the vehicles in each input image, producing a form of IR. 
In contrast to the works mentioned above, \textsc{roadscene2vec} utilizes a \textit{scene-graph} IR that encodes the spatial and semantic relations between all the traffic participants in a frame. This form of representation is similar to a knowledge graph with the key distinction that \textit{scene-graphs} explicitly encode knowledge about a visual scene.



\subsection{Graph-based Driving Scene Understanding}
\textcolor{green}{Several works have applied graph-based formulations for road scene understanding. 
In~\cite{li2019learning}, the authors propose a 3D-aware egocentric Spatio-temporal interaction framework that uses both an \textit{Ego-Thing} graph and an \textit{Ego-Stuff} graph to encode how the ego vehicle interacts with both moving and stationary objects in a scene, respectively.
In~\cite{mylavarapu2020towards}, the authors propose a pipeline using a multi-relational graph convolutional network (MR-GCN) for classifying the driving behaviors of traffic participants.
The MR-GCN combines spatial and temporal information, including relational information between moving and landmark objects. 
In \cite{tian2020road}, the authors propose extracting road scene graphs in a manner that includes pose information for scene layout reconstruction. A similar approach was also proposed in \cite{kunze2018reading}. 
Authors in \cite{liu2021real} propose using a probabilistic graph approach for explainable traffic collision inference.
In our prior work, we demonstrated that a \textit{scene-graph} representation used with an MRGCN leads to state-of-the-art performance at assessing the subjective risk of driving maneuvers \cite{yu2021scene}.
Our tool implements examples of multi-relational graph learning models (MRGCN and MRGIN) and model skeletons, enabling users to evaluate other graph learning model formulations more easily.}

\subsection{Graph Extraction and Graph Learning Libraries}
\textcolor{green}{Other libraries for extracting \textit{scene-graphs} from input images have been proposed. \cite{yang2018graph} proposed the Graph R-CNN model, which extracts scene graphs by identifying the set of individual objects in the image before identifying the spatial relations between the objects. With this process, Graph R-CNN can extract the spatial features of the scene in the form of a \textit{scene-graph}. \cite{tang2020sgbenchmark} provides a benchmark for evaluating several kinds of \textit{scene-graph} generation models on image datasets. The \textit{scene-graph} representations extracted by these tools are then used for semantic understanding and labeling tasks, such as image captioning and visual question answering. Although these tools and models are successful at these tasks, they do not incorporate specific domain knowledge relevant to the AV problem space. Autonomous driving is a highly complex problem on its own, so AV algorithms must utilize domain knowledge, including driving rules, road layout, and markings, as well as light and sign information. Furthermore, AV algorithms must account for temporal factors; the tools mentioned above operate on individual images and thus do not account for these safety-critical temporal factors.}

Regarding graph learning tools and libraries, several tools such as GraphGYM \cite{you2020design}, DGL \cite{wang2019deep}, and OGB \cite{hu2020open} exist for quickly and easily evaluating several graph learning models on problems including node/graph classification and regression. However, none of these pre-existing tools enable \textit{scene-graph} generation; they can only be used with existing graph data. \textcolor{green}{Our proposed tool is the only tool that enables both the extraction and learning of AV-specific \textit{scene-graphs}.} 
\section{\textsc{Roadscene2vec} Architecture}
\label{sec:roadscene2vec}
This section introduces \textsc{roadscene2vec}'s architecture, features, and intended workflow. Our \textsc{roadscene2vec} is implemented as a Python library, integrating various external packages such as APIs from PyTorch, PyTorch Geometric, Detectron2, and CARLA.
\textsc{roadscene2vec} consists of four key modules: (i) data generation (\textbf{data.gen}) and preprocessing (\textbf{data.proc}), (ii) \textit{scene-graph} extraction  (\textbf{scene\_graph}), (iii) model training and evaluation (\textbf{learning}), and (iv) visualization (\textbf{util}). We detail each module in the following subsections. 


\begin{figure}[htb]
    \centering
    \includegraphics[clip, trim=0 287 820 0, width=1.0\linewidth]{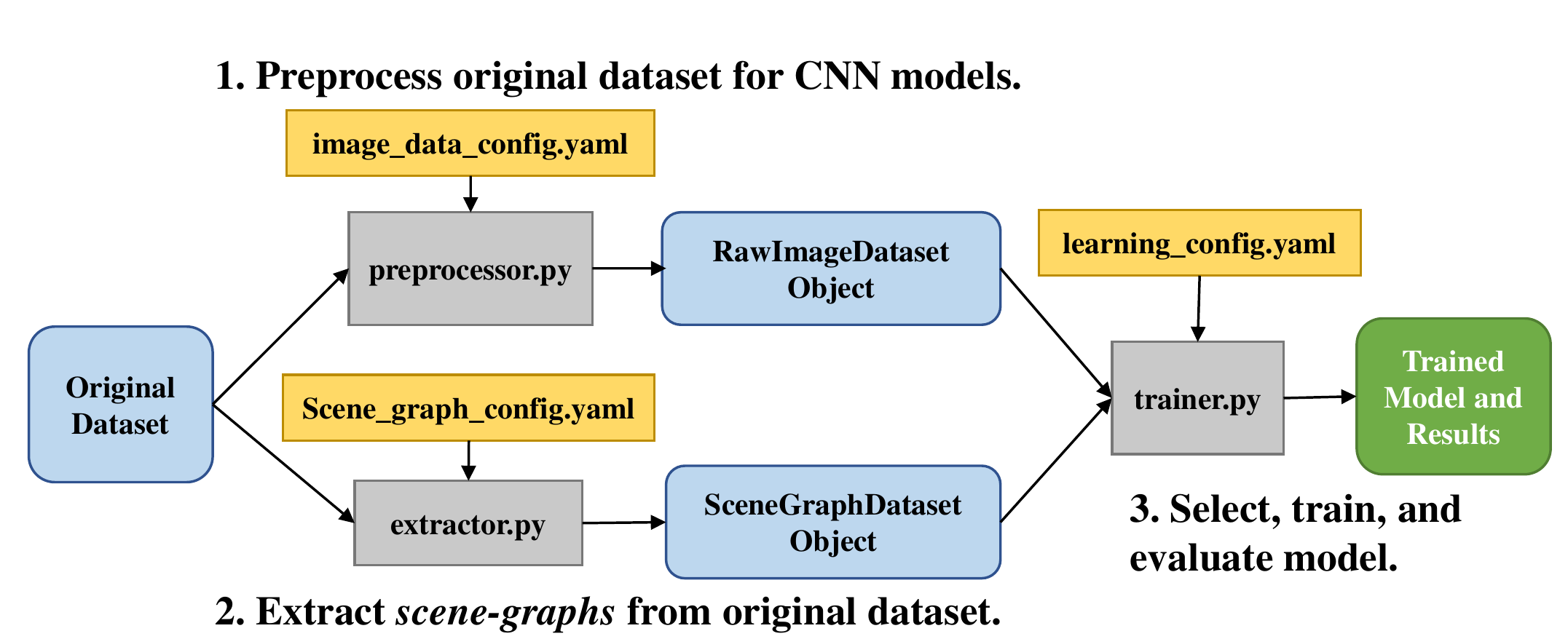}
    \caption{Workflow for using \textsc{roadscene2vec} to preprocess a dataset; extract \textit{scene-graphs} from the dataset; and select, train, and evaluate a model on the dataset.}
    \label{fig:roadscene2vec_workflow}
\end{figure}

\subsection{Dataset Generation Tools (data.gen)}
\label{subsec:data.gen}
The module \textbf{data.gen} in \textsc{roadscene2vec} allow researchers to synthesize driving data for their research.
To successfully handle complex and long-tail driving scenarios, deep learning approaches typically train their models on large datasets that contain a wide range of "corner cases."
However, generating such datasets is expensive and time-consuming in the real-world~\cite{dosovitskiy2017carla}. 
Thus, most researchers instead use synthesized datasets containing plenty of these corner cases to evaluate their research ideas.

For this purpose, \textsc{roadscene2vec} integrates the open-source driving simulator, CARLA~\cite{dosovitskiy2017carla}, which allows users to generate driving data by controlling a vehicle (either in manual mode or autopilot mode) in simulated driving scenarios. On top of that, \textsc{roadscene2vec} also integrates the CARLA Scenario Runner, which contains a set of atomic controllers that enable users to automate the execution of complex driving maneuvers.

In \textsc{roadscene2vec}, \textbf{data.gen} produces each driving clip in CARLA's simulated world by (i) selecting one autonomous car randomly, (ii) switching its mode to manual mode, and (iii) using the Scenario Runner to command the vehicle to change lanes. 
In addition, the data generation tool in \textsc{roadscene2vec} manipulates the various presets in CARLA to specify the number of cars, pedestrians, weather and lighting conditions, etc., for making the generated driving data more diverse.
Moreover, through the APIs provided by the Traffic Manager (TM) of the CARLA simulator, the tool can customize the driving characteristics of every autonomous vehicle in the simulated world, such as the intended speed considering the current speed limit, the chance of ignoring the traffic lights, or the chance of neglecting collisions with other vehicles.
Overall, the tool allows users to simulate a wide range of very realistic urban driving environments and generate synthesized datasets suitable for training and testing a model.

Using the CARLA Python API and the CARLA Scenario Runner, we implemented a tool in the \textbf{data.gen} module for extracting the road scene's state information as well as the corresponding ego-centric camera images directly from the CARLA simulator for use in \textsc{roadscene2vec}. 
For each frame in a driving sequence, we store the attributes of all the objects as a Python dictionary. These attributes include object type, location, rotation, lane assignment, acceleration, velocity, and light status. For static objects such as traffic lights and signage, we store the type of object, its location, and light state (light color) or sign value (e.g., speed limit). We refer to the datasets in this format as CARLA datasets. In addition, our tool supports using image-based datasets, such as the camera data extracted from CARLA or the Honda Driving Dataset \cite{Ramanishka_behavior_CVPR_2018} used in our experiments.
The code provided in our data.gen module can be modified to support other driving actions, such as turning, accelerating, braking, and overtaking. 

\begin{figure}[htb]
    \centering
    \includegraphics[width=.8\linewidth]{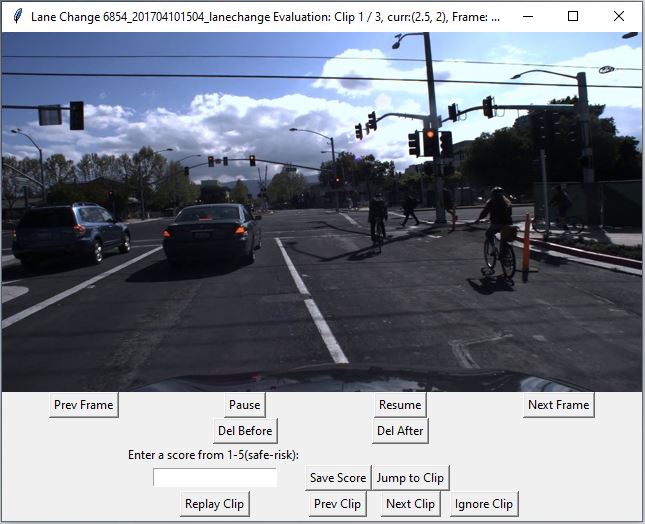}
    \caption{The user interface of the annotator tool, used to label, filter, and trim datasets.}
    \label{fig:annotator}
\end{figure}

Under the \textbf{data.gen} module, \textsc{roadscene2vec} also provides an annotation tool for quickly and easily labeling both CARLA datasets and image datasets. The annotator offers a graphical user interface (GUI) that enables users to view, label, exclude or trim specific driving sequences. 
Our annotator enables users to assign one label for each sequence and supports averaging multiple independent labelers' decisions. Our annotators GUI is shown in Figure \ref{fig:annotator}. 
\green{In comparison to popular annotation tools, such as CVAT \cite{sekachev2020opencv} and VoTT \cite{microsoft2021Dec}, our annotator offers a more streamlined approach for video clip labeling. These other annotation tools typically iterate through only a single image at a time as they are primarily designed for tasks such as object detection and semantic segmentation. In contrast, our annotator implements a broader range of video playback controls (play, pause, replay, ignore, etc.), facilitating risk analysis over a complete video sequence. Currently, our annotator supports sequence-level risk labels from 1-5. Still, it can be extended to support other label types for problems such as scenario classification and rare event identification.}
In addition to the annotation tool, we also provide dataset utilities such as train-test splitting, k-fold cross-validation, and downsampling as part of the trainers in the \textbf{learning.util} module.

\subsection{Data Preprocessing (roadscene2vec.data.proc)}
The data storage and preprocessing functions are implemented through the \textbf{data.proc} module of \textsc{roadscene2vec}.
To use a new dataset with \textsc{roadscene2vec}, it must first have the correct directory structure defined in our repository. Next, the input dataset can go through one of the two workflows shown in Figure \ref{fig:roadscene2vec_workflow}: (i) the dataset is preprocessed into a "RawImageDataset" to be used with CNNs and other image processing models directly, or (ii) the dataset is sent to the corresponding \textit{scene-graph} extractor to generate \textit{scene-graph} representations of every frame in the dataset (discussed in Section \ref{subsec:extraction}).
The preprocessing step is necessary for the conventional deep-learning models; the input images often need to be resized, reshaped, or sub-sampled before being trained with models to meet memory and space constraints. 
After preprocessing, the RawImageDataset object stores the sets of driving video clips as image sequences, the labels associated with the video clips, and metadata (such as sequence name/action type). 
For each image in each clip in the dataset, the image preprocessor loads the image using OpenCV, resizes and recolors the image according to the configuration settings, and stores the image as a PyTorch Tensor. The resulting RawImageDataset object is then serialized and stored as a pickle (.pkl) file.

\subsection{Road \textit{Scene-Graph} Extraction (roadscene2vec.scene\_graph.extraction)}
\label{subsec:extraction}
Here, we describe how an input dataset is converted into a "SceneGraphDataset" object via our \textit{scene-graph} extraction framework. We first describe how the entities and relations in the \textit{scene-graph} are defined and configured before discussing the specific steps needed to extract \textit{scene-graphs} from both CARLA and image-based datasets.

\subsubsection{Entity and Relation Extraction}
\label{subsec:relation_extraction}

\begin{table}[htb]
    \centering
    \begin{tabular}{p{120pt} p{200pt}}
    \hline
Parameter & Description \\\hline
actor\_names & The list of object types. The default list is based on the actor types defined by the CARLA simulator.\\
relation\_names & The list of all implemented relation types. \\
car\_names / moto\_names / bicycle\_names / etc. & Object names defined in the CARLA simulator. These lists are used to cross-reference the object type for a given CARLA vehicle name.\\\hline

directional\_thresholds & Defines the set of enabled directional relations and their thresholds in degrees.\\
directional\_relation\_list & Defines the pairs of object types for which directional relations will be extracted.\\
proximity\_thresholds & Defines the set of enabled distance relations and their thresholds in feet.\\
proximity\_relation\_list & Defines the pairs of object types for which proximity relations will be extracted.\\

lane\_threshold & Represents 50\% of the width of a lane in feet. If an object is more than this distance from the ego car's center, it is considered to be in the left or right lane.\\
\hline
    \end{tabular}
    \caption{Scene graph configuration options and their descriptions. Each of these parameters can be reconfigured by the user to produce custom \textit{scene-graphs}.}
    \label{tab:extraction_parameters}
\end{table}

A list of \textsc{roadscene2vec}'s user-configurable \textit{scene-graph} extraction settings is shown in Table \ref{tab:extraction_parameters}. In our formulation, each "actor" (object) in the \textit{scene-graph} is assigned a type from the set \{car, motorcycle, bicycle, pedestrian, lane, light, sign\}, matching those defined by CARLA. Users can reconfigure the set of object types to support other dataset types, applications, or ontologies. 

The default relation extraction pipeline we implement identifies three kinds of pair-wise relations: \textit{proximity} relations (e.g. \textit{visible}, \textit{near}, \textit{very\_near}, etc.), \textit{directional} relations (e.g. \textit{Front\_Left}, \textit{Rear\_Right}, etc.), and \textit{belonging} relations (e.g. car\_1 \textit{isIn} left\_lane). 
Two objects are assigned the \textit{proximity} relation, $r\in$ \{\textit{Near\_Collision} (4 ft.), \textit{Super\_Near} (7 ft.), \textit{Very\_Near} (10 ft.), \textit{Near} (16 ft.), \textit{Visible} (25 ft.)\} provided the objects are physically separated by a distance that is within that relation's threshold. 
The {\it directional relation}, $r \in$ \{\textit{Front\_Left}, \textit{Left\_Front}, \textit{Left\_Rear}, \textit{Rear\_Left}, \textit{Rear\_Right}, \textit{Right\_Rear}, \textit{Right\_Front}, \textit{Front\_Right}\}, is assigned to a pair of objects, in this case between the ego-car and another car in the view, based on their relative orientation and only if 
they are within the \textit{near} threshold distance from one another. 
Additionally, the \textit{isIn} relation identifies which vehicles are on which lanes (see Fig.~\ref{fig:scene_graph}). We use each vehicle's horizontal displacement relative to the ego vehicle to assign vehicles to either the \textit{Left Lane}, \textit{Middle Lane}, or \textit{Right Lane} using the known lane width. 
Our current abstraction only considers three-lane areas, and, as such, we map vehicles in all left lanes and all right lanes to the same \textit{Left Lane} node \textit{Right Lane} node, respectively. 
If a vehicle overlaps two lanes (i.e., during a lane change), it is mapped to both lanes.

The set of possible entity types, relation types, relation thresholds, and valid object pairs is defined in the \textit{scene\_graph\_config} file. These settings are entirely user re-configurable, enabling broad design space exploration of different graph extraction methodologies. 
After graph extraction is completed, the set of all \textit{scene-graph} sequences, metadata, and labels are saved as a SceneGraphDataset.

\subsubsection{CARLA \textit{Scene-Graph} Extraction}
Since the CARLA datasets contain a dictionary with a list of objects and their attributes, we directly use this dictionary to initialize the nodes in the \textit{scene-graph}. Each node is assigned its type label from the set of actor\_names and its corresponding attributes (e.g., position, angle, velocity, current lane, light status, etc.) for relation extraction. Once all nodes are added to the \textit{scene-graph}, we extract relations between each pair of objects in the scene.  

\subsubsection{Image \textit{Scene-Graph} Extraction}
\green{To extract \textit{scene-graphs} from image-based datasets, we first need to identify the set of objects in each image along with their attributes}. We use Mask-RCNN \cite{he2017mask} to extract the set of objects in the image as well as their bounding boxes. Next, we compute the inverse-perspective mapping transformation of the image, yielding a top-down 'birds-eye view' (BEV) projection of the scene. By generating this projection and projecting the bounding box coordinates from the original image into the birds-eye view, we can estimate the position of each vehicle relative to the ego-vehicle with reasonably high fidelity. This position information, along with the object class information, is used to construct the \textit{scene-graphs}. 
However, the BEV projection needs to be re-calibrated for each dataset, as typically, each dataset uses a different camera angle and camera configuration. To facilitate this calibration step, we provide a BEV calibration utility in \textbf{scene\_graph.extraction.bev}. This utility provides an interactive way for the user to select the road area and calibrate the BEV projection for a new dataset with a single step.

\subsubsection{\textit{Scene-Graph} Visualization}
Our \textit{scene-graph} visualization tool, located in the  \textbf{roadscene2vec.util} module, consists of a GUI that simultaneously displays an input image side by side with its corresponding \textit{scene-graph}, as is shown in Figure \ref{fig:visualizer}. 
This tool enables researchers to experiment with a wide range of relation types and distance thresholds and quickly optimize their \textit{scene-graph} extraction settings for their specific application or dataset. 

\subsection{\textit{Scene-Graph} Embedding (roadscene2vec.learning)}
\green{The \textbf{learning} module contains our framework for splitting datasets as well as training, testing, and scoring models at various tasks. It also contains our graph learning models, baseline deep learning models, and a graph learning model template to enable users to define their own graph models for use with our framework.}. The \textbf{model} submodule contains the model definitions while the \textbf{util} submodule contains the training, evaluation, and scoring functions. 
The training code supports implementing k-fold cross-validation, a user-definable train-test split, and downsampling and class weighting to correct dataset imbalances. 
The model specification, training hyperparameters, and dataset configuration settings are loaded from the \textit{learning\_config} file, which is user-modifiable. Next, we introduce the models available in \textsc{roadscene2vec}.

\subsubsection{Graph Learning Models (roadscene2vec.learning.model)}
\label{subsec:graph_learning}
\begin{figure}[ht]
    \centering
    \includegraphics[clip, trim= 10 148 450 29, width=\linewidth]{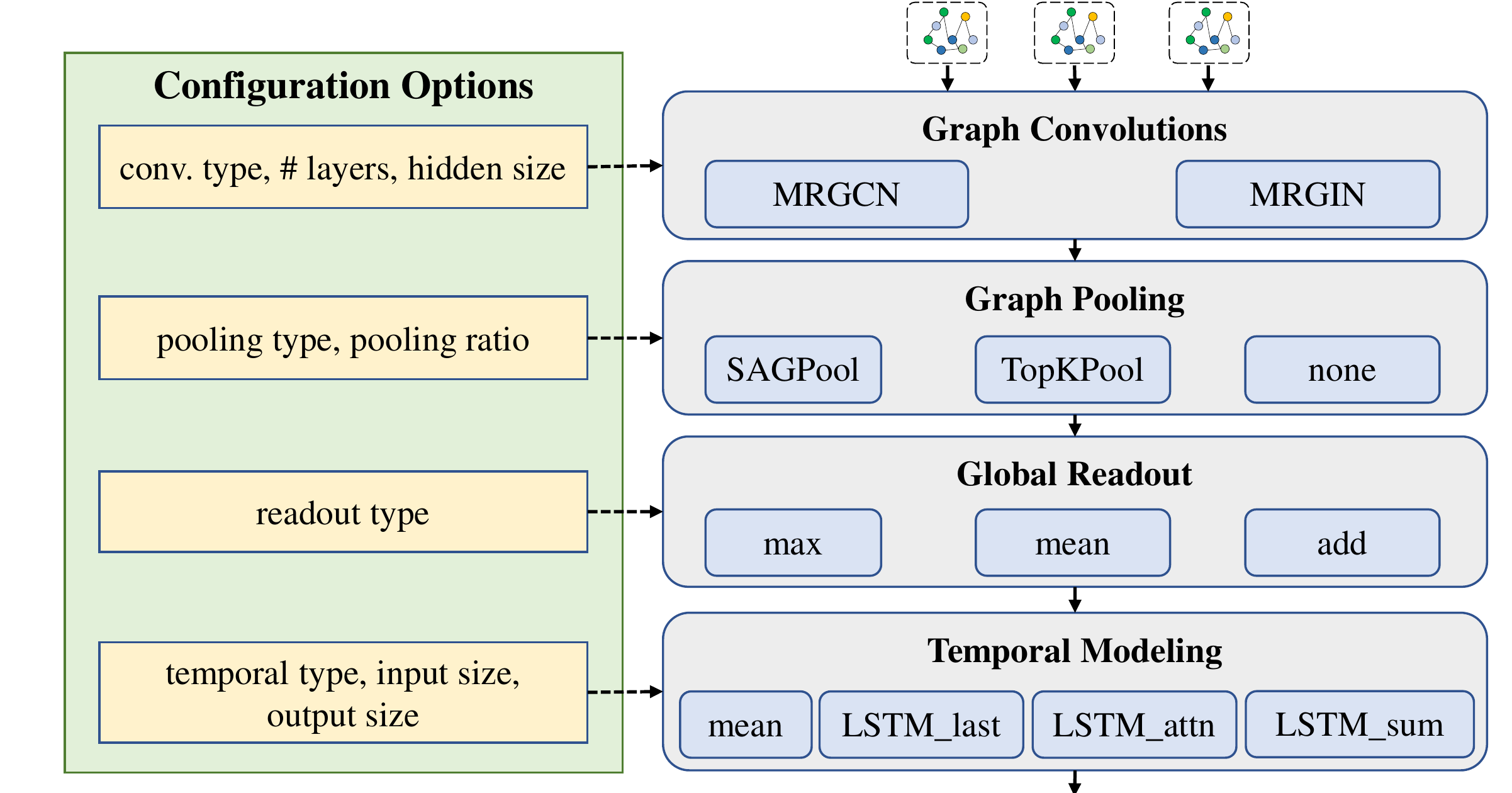}
    \caption{Graph learning model configuration options provided in \textsc{roadscene2vec}.}
    \label{fig:graph_learning}
\end{figure}
The graph learning models we provide in \textsc{roadscene2vec} enable various configurations of both spatial modeling and temporal modeling components as shown in Figure \ref{fig:graph_learning}.
The spatial modeling components that are configurable include
(i) graph convolution layers, (ii) graph pooling and graph attention layers, and (iii) graph readout operations. The configurable temporal modeling components include (i) temporal modeling layers and (ii) temporal attention layers. 
Our experiments use MRGCN and MRGIN models that are identical in structure and differ only in the type of spatial modeling used.  
Next, we discuss these components in more detail.

\paragraph{Spatial Modeling \textsc{(Spatial\_Model)}}
We provide two multi-relational graph convolution implementations based on (i) graph convolutional networks (GCNs) \cite{kipf2016semi} and (ii) graph isomorphism networks (GINs) \cite{xu2018powerful}. These layers propagate node embeddings across edges via graph convolutions, resulting in a new set of node embeddings. \green{The two implementations differ in how data is propagated through successive graph convolutions. }
Graph pooling filters the set of node embeddings in the graph to only those most useful for the task. 
We enable two types of graph pooling layers extended for multi-relational use cases: Self-Attention Graph Pooling (SAGPool)~\cite{lee2019self} and Top-K Pooling (TopkPool)~\cite{gao2019graph}. 
After pooling, a global readout operation is used to collect the set of pooled node embeddings into a unified graph embedding. We implement \textit{max}, \textit{mean}, and \textit{add} readout operations.

\paragraph{Temporal Modeling \textsc{(Temporal\_Model)}}
The temporal model we implement uses Long Short-Term Memory (LSTM) layers to convert the sequence of \textit{scene-graph} embeddings to either (i) one spatio-temporal embedding (for sequence classification tasks) or (ii) a sequence of spatio-temporal embeddings (for graph classification/prediction tasks). For graph classification/collision prediction tasks, the output from an LSTM layer for each input \textit{scene-graph} embedding is collected as a sequence of spatio-temporal \textit{scene-graph} embeddings $P$ that is sent to an MLP layer to produce the final set of model outputs.
For sequence classification tasks, a temporal readout operation is applied to $P$ to compute a single spatio-temporal sequence embedding $\mathbf{z}$
by (i) extracting only the last hidden state of the LSTM $p_T$ (LSTM-last), (ii) taking the sum over $\mathbf{P}$, or (iii) using a temporal attention layer (LSTM-attn) to compute an attention-weighted sum of the different elements of $P$ as described in \cite{bahdanau2014neural}.

\subsubsection{Baseline Models (roadscene2vec.learning.model)}
In addition to the graph learning models that are core to \textsc{roadscene2vec}, we also provide a set of baseline deep learning models for quickly and easily comparing to typical image-processing approaches. These baselines include (i) a ResNet-50 \cite{he2016deep} CNN classifier and (ii) a CNN+LSTM classifier \cite{yurtsever2019risky}. The motivation for using these baselines stems from their prevalence in AV image processing tasks, such as risk assessment \cite{yurtsever2019risky}. Users can easily use other graph or deep-learning models with our framework as long as they follow the same, typical PyTorch model structure.

\subsubsection{Performance Evaluation and Hyperparameter Optimization}
To enable live monitoring of training runs and in-depth analysis of the effects of different hyperparameter settings on performance, we integrate our library with Weights and Biases (W\&B)\footnote{\url{https://wandb.ai/}}. W\&B is a free, publicly available tool for tracking experiments, visualizing performance, identifying hyperparameter importance, and organizing results. 
We believe this integration will enable researchers to identify trends in the data and optimize model performance more quickly.

\section{Usage Examples}
\label{sec:usage}
In this section, we describe some of \textsc{roadscene2vec}'s use-cases.
First, Section~\ref{sec:use-case-1} exhibits a fundamental use-case in which an image frame $I$ is converted into a \textit{scene-graph} $g$ and then into a fixed-length embedding $h_g$. 
Next, the use cases of \textsc{roadscene2vec} for two risk-based autonomous driving applications (subjective risk assessment and collision prediction) are described in Section~\ref{sec:use-case-2} and Section~\ref{sec:use-case-3}, respectively. In Section \ref{subsec:use-case-4}, we discuss how \textsc{roadscene2vec} can be used for performing and evaluating transfer learning. Finally, in Section \ref{subsec:use-case-5}, we show how \textsc{roadscene2vec} can be used to analyze the explainability of the graph learning models. 

\subsection{Use Case 1: Converting an Ego-Centric Observation Into a Scene-Graph}
\label{sec:use-case-1}
Our high-level algorithm for converting an input image into a \textit{scene-graph} is shown in Algorithm \ref{alg:use-case-1}. 
Let us walk through a typical workflow for converting an image dataset into a set of \textit{scene-graph} embeddings.
\textcolor{green}{First, the preprocessor processes the image to set the dataset format and image sizing.} These \textit{scene-graphs} can then be visualized using the visualizer tool we provide. The following script streamlines the execution of this use case:
\begin{verbatim}
    > python examples/use_case_1.py
\end{verbatim}
These scripts take configuration information directly from the data\_config and scene\_graph\_config files in the \textbf{config} module. The config files indicate which type of dataset is being used (CARLA or image-based) and the location and extraction settings for the dataset. The scene\_graph\_config file also allows the reconfiguration of the relation extraction settings as shown in Table \ref{tab:extraction_parameters}. 
The choice of relation extraction settings changes the \textit{scene-graph} structure, which can change how the graph learning model processes the data.

\begin{algorithm}[ht]
    \SetAlgoLined
    \DontPrintSemicolon
    \textbf{Input:} A sequence of images from a driving video clip $I$.\;
    \textbf{Output:} A sequence of scene graphs $G$ for $I$.\;
    \SetKwProg{Fn}{def}{:}{}
    \SetKwFunction{Fhwgraph}{\textit{IMG2GRAPH}}
    \SetKwFunction{Fgraphvec}{\textit{EXTRACT\_SEQ}}
    \Fn{\Fhwgraph{$I_t$}}{
        $O_t \gets$ \textsc{Obj\_Detection}($I_t$)\;
        $A_t \gets$ \textsc{Attr\_Extraction}($I_t, O_t$)\;
        $G_t \gets$ \textsc{Graph\_Extraction}($O_t, A_t$)\;
        \KwRet $G_t$\;
    }
    \Fn{\Fgraphvec{$I$}}{
        $G \gets$ \{ \}\;
        \For{$I_t$ in $I$}{
        $G_t \gets$ \textit{\texttt{IMG2GRAPH}}($I_t$)\;}
        \KwRet $G$\;
    }
    $G \gets$ \textit{\texttt{EXTRACT\_SEQ}}($I$)\;
    \caption{Use Case 1 - Extracting a sequence of \textit{scene-graphs} from a driving clip.}
    \label{alg:use-case-1}
\end{algorithm}

\subsection{Use Case 2: Subjective Risk Assessment}
\label{sec:use-case-2} 

\begin{figure}[htb]
    \centering
    \includegraphics[clip, trim=104 136 50 155,width=\linewidth]{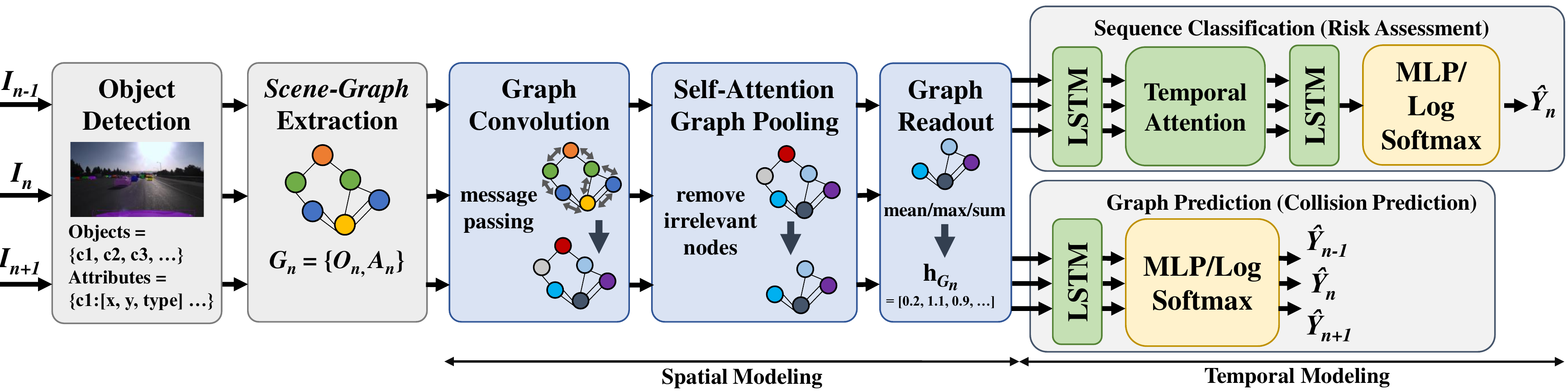}
    \caption{The architecture of our configurable \textit{scene-graph} based AV perception model. Our two pre-implemented temporal modeling pipelines for specific AV tasks are shown (sequence classification and graph prediction). However, users can remove or replace these model components for performing other AV tasks such as graph classification or scenario classification.}
    \label{fig:archi}
\end{figure}
In prior AV research, attempts to improve vehicle safety have involved modeling either the \textit{objective risk} or the \textit{subjective risk} of driving scenes~\cite{grayson2003risk, fuller2005towards, bao2019personalized}.
The \textit{objective risk} is defined as the objective probability of an accident occurring and is typically determined by statistical analysis~\cite{grayson2003risk}. 
In contrast, \textit{subjective risk} refers to the driver's own perceived risk and is an output of the driver's cognitive process~\cite{fuller2005towards, bao2019personalized}.
Since subjective risk accounts for the human behavior perspective and its critical role in anticipating risks \cite{bao2020personalized, bao2019personalized,fuller2005towards}, it has the potential to assess contextual risk better than objective methods and thus better assure passenger safety.
Further, studies such as ~\cite{trankle1990risk,grayson2003risk} provide direct evidence that a driver's subjective risk assessment is inversely related to the risk of traffic accidents.
Within this context, AVs must \textcolor{green}{support understanding} driving scenes and quantify the subjective risk of driving decisions.

Given this motivation, we show that the graph learning models available in \textsc{roadscene2vec} can be used to convert these extracted \textit{scene-graphs} into spatio-temporal \textit{scene-graph} embeddings for the task of subjective risk assessment, as was done in our prior work \cite{yu2021scene}. 

\subsubsection{Problem Formulation}
In our prior work \cite{yu2021scene}, and here, we make the same assumption used in~\cite{yurtsever2019risky} that the set of driving sequences can be partitioned into two jointly exhaustive and mutually exclusive subsets: risky and safe. 
We denote the sequence of images of length $T$ by $\mathbf{I} = \{I_1, I_{2}, I_{3}, ..., I_T\}$.
We assume the existence of a spatio-temporal function $f$ that outputs whether a sequence of driving actions $x$ is safe or risky via a risk label $y$, as given in Equation~\ref{formular1-v2}.
\begin{equation}
\label{formular1-v2}
    y = f(\mathbf{I}) = f(\{I_1, I_2, I_3, ..., I_{T-1}, I_T\}),
\end{equation} 
where 
\begin{equation}
    y=\left \{
    \begin{array}{ll}
         (1,0), &  \text{if the driving sequence is safe}\\
         (0,1), &  \text{if the driving sequence is risky}.
    \end{array}
    \right.
\end{equation}

Overall, the goal of the model is to learn to approximate the function $f$. Our algorithmic implementation of this use case is shown in Algorithm \ref{alg:riskassessment}.

\begin{algorithm}[ht]
\SetAlgoLined
\LinesNumbered
\DontPrintSemicolon
\textbf{Input: A sequence of images from a driving video clip $I$.}\;
\textbf{Output: Risk assessment $\hat{Y}$.}\;
\SetKwFunction{Fsgvec}{SEQ2VEC}
\SetKwFunction{FMain}{RISK\_ASSESS}
\SetKwProg{Fn}{def}{:}{}
\Fn{\Fsgvec{$G$}}{
$h_G \gets$ \{ \}\;
\For{$G_t$ in $G$}{
$\mathbf{h}_{G_t} \gets$ \textsc{Spatial\_Model}($G_{t}$)\;}
$Z \gets$ \textsc{Temporal\_Model}($\mathbf{h}_{G}$)\;
$\hat{y}_0, \hat{y}_1 \gets$ \textsc{Activation}(MLP($Z$))\;
\uIf{$\hat{y}_1 \geq \hat{y}_0$}{ 
\KwRet $1$\;}
\uElseIf{$\hat{y}_0 > \hat{y}_1$}{
\KwRet $0$\;}
}
\Fn{\FMain{$I$}}{
$G \gets$ EXTRACT\_SEQ($I$)\;
$\hat{Y}\gets$ SEQ2VEC($G$)\;
\KwRet{$\hat{Y}$}\;
}
$\hat{Y} \gets$ RISK\_ASSESS($I$)\;
\caption{Use Case 2 - \textit{Scene-graph} embedding for risk assessment}
\label{alg:riskassessment}
\end{algorithm}

\subsubsection{Training}
To achieve this goal, we train the graph learning model using the extracted sequences of \textit{scene-graphs} as inputs and the subjective risk labels given by human annotators for each sequence. As such, the problem becomes a simple sequence classification problem, where the goal is to classify a given sequence of images as \textit{risky} or \textit{safe}. The configuration settings for training the model are available in the learning\_config file in the \textbf{config} module.
The following command can be used to train the model for risk assessment:
\begin{verbatim}
    > python examples/use_case_2.py
\end{verbatim}

\subsection{Use Case 3: Collision Prediction}
\label{sec:use-case-3}
\begin{figure}[ht]
    \centering
    \includegraphics[width=\linewidth]{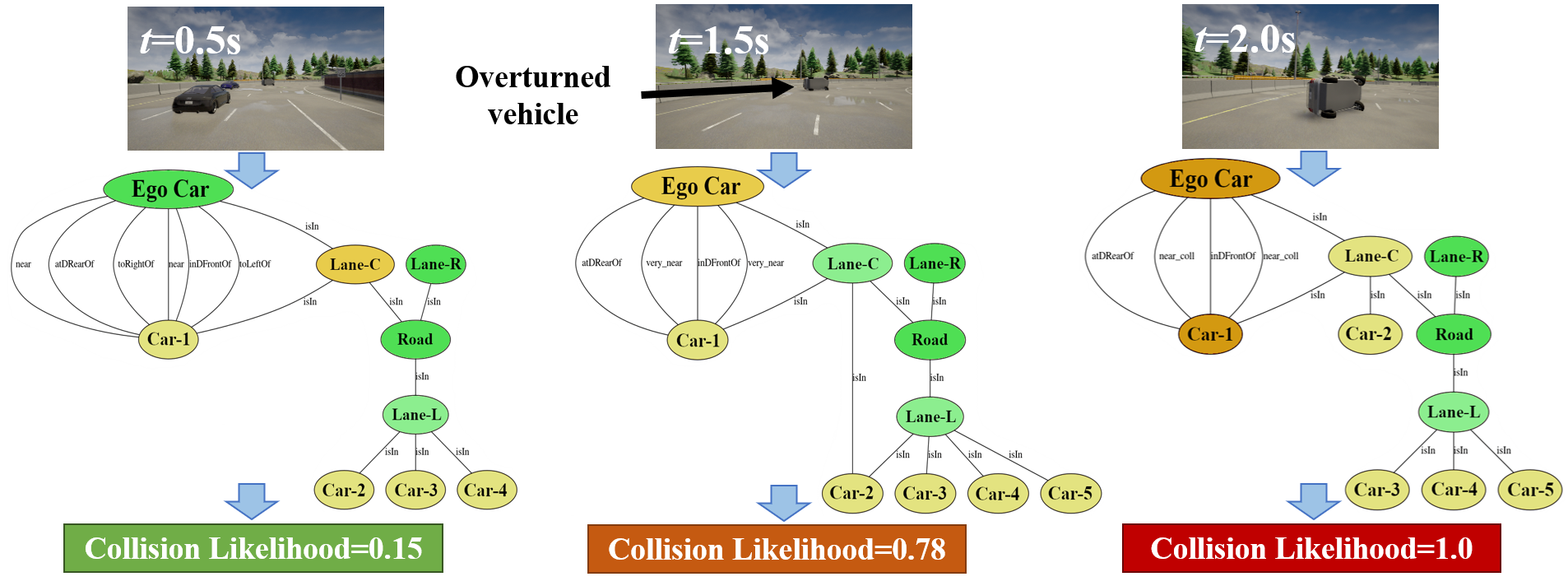}
    \caption{Demonstration of collision prediction using \textit{scene-graphs}. Each node's color indicates its attention score (importance to the collision likelihood) from orange (high) to green (low).}
    \label{fig:usecase3}
\end{figure}

In our third use case, we demonstrate how \textsc{roadscene2vec} can be used to study approaches for predicting future vehicle collisions.
In contrast to Use Case 2, which is a sequence classification problem, collision prediction has safety-critical time constraints. 
\textcolor{green}{It uses the history of prior \textit{scene-graphs} to make predictions about the state of future graphs.}
Current statistics indicate that perception and prediction errors were factors in over 40\% of driver-related crashes between conventional vehicles \cite{mueller2020humanlike}. 
However, a significant number of reported AV collisions are also the result of these errors \cite{schoettle2015preliminary, xu2019statistical}. With this motivation, we show that \textit{scene-graphs} can be used to represent road scenes and model inter-object relationships to improve perception and scene understanding. An example of our methodology is shown in Figure \ref{fig:usecase3}.

\subsubsection{Problem Formulation}
We formulate the problem of collision prediction as a time-series classification problem where the goal is to predict if a collision will occur in the \textit{near future}. 
Our goal is to accurately model the spatio-temporal function $f$, where
\begin{equation}
    \mathbf{Y_n} = f(\{I_{1}, ..., I_{n-1}, I_n\}), \mathbf{Y_n} \in \{0,1\}, \text{for } n > 2,
\end{equation}
where $\mathbf{Y_n}=1$ implies a collision in the near future and $\mathbf{Y_n} = 0$ otherwise.
Here the variable $I_n$ denotes the image captured by the on-board camera at time $n$. The interval between each frame varies with the camera sampling rate. Our implementation of Use Case 3 is shown in Algorithm \ref{alg:riskassessment}.

\begin{algorithm}[ht]
\SetAlgoLined
\LinesNumbered
\DontPrintSemicolon
\textbf{Input: A sequence of images from a driving video clip $I$.}\;
\textbf{Output: Sequence of collision likelihood predictions: $\hat{Y}$.}\;
\SetKwFunction{Fsgvec}{GRAPH2VEC}
\SetKwFunction{FMain}{COLLISION\_PRED}
\SetKwProg{Fn}{def}{:}{}
\Fn{\Fsgvec{$G_t$, $p_{t-1}$, $c_{t-1}$}}{
$\mathbf{h}_{G_t} \gets$ \textsc{Spatial\_Model}($G_{t}$)\;
$p_t, c_{t} \gets$ \textsc{Temporal\_Model}($\mathbf{h}_{G_t}, p_{t-1}, c_{t-1}$)\;
$\hat{y}_0, \hat{y}_1 \gets$ \textsc{Activation}(MLP($p_t$))\;
\uIf{$\hat{y}_1 \geq \hat{y}_0$}{ 
\KwRet $1, p_{t}$\;}
\uElseIf{$\hat{y}_0 > \hat{y}_1$}{
\KwRet $0, p_{t}$\;}
}
\Fn{\FMain{$I$}}{
$G \gets$ EXTRACT\_SEQ($I$)\;
$p_0, c_0 \gets $ [0, 0, ..., 0] , [0, 0, ..., 0]\;
$\hat{Y} \gets$ \{ \}\; 
\For{$G_t$ in $G$}{
$\hat{Y}_t, p_t \gets$ GRAPH2VEC($G_t$, $p_{t-1}$, $c_{t-1}$)\;
$t \gets t+1$\;}
\KwRet{$\hat{Y}$}\;
}
$\hat{Y} \gets$ COLLISION\_PRED($I$)\;
\caption{Use Case 3 - \textit{Scene-graph} embedding for collision prediction}
\label{alg:collprediction}
\end{algorithm}

\subsubsection{Training}
To train a model for this application, we adjust the model to produce one output per graph instead of one output per sequence. For the application of collision prediction, we also assign each frame in a video clip a label identical to the entire clip's label to train the model to identify the preconditions of a future collision and predict it as early as possible. The following command can be used to train the model for collision prediction:
\begin{verbatim}
    > python examples/use_case_3.py
\end{verbatim}
 
\subsection{Use Case 4: Transfer Learning}
\label{subsec:use-case-4}
Models trained on simulated datasets must be able to transfer their knowledge to real-world driving scenarios as they can differ significantly from simulations. 
One key advantage of using \textit{scene-graphs} is that they are a form of Intermediate Representation (IR), meaning that they provide a higher level of abstraction compared to image data alone. This abstraction means that \textit{scene-graphs} are generally better able to transfer knowledge across datasets and domains, such as from simulated data to real-world driving data. 
Since this is a key benefit of using a graph-based approach and is a critical use case for validating AV safety, \textsc{roadscene2vec} supports running transfer learning experiments between any two datasets. To implement this use case, we use the original dataset to train the model and use the user-specified transfer dataset to test the model. No additional domain adaptation is performed. The workflow for Use Case 4 is shown in Algorithm \ref{alg:transfer}.
The following script runs an example of transfer learning.
\begin{verbatim}
    > python examples/use_case_4.py
\end{verbatim}

\begin{algorithm}[ht]
\SetAlgoLined
\LinesNumbered
\DontPrintSemicolon
\textbf{Input: Source dataset $D_S$, transfer dataset $D_T$, model $m$, and training epochs $E$.}\;
\textbf{Output: Transfer learning result $R_T$.}\;
\SetKwFunction{Fsgvec}{TRAIN}
\SetKwFunction{Feval}{EVALUATE}
\SetKwFunction{FMain}{TRANSFER\_KNOWLEDGE}
\SetKwProg{Fn}{def}{:}{}
\Fn{\Fsgvec{$D$, $m$, $E$}}{
\For{$epoch$ in $E$}{
$X, Y \gets D$\;
$O \gets m(X)$\;
$L \gets$ \textsc{Loss\_Function}($O$, $Y$)\;
$m \gets$ \textsc{Update\_Model}($L$, $m$)\;
}
\KwRet{$m$}
}
\Fn{\Feval{$D$, $m$}}{
$X, Y \gets D$\;
$O \gets m(X)$\;
$R \gets$ \textsc{Score}($O$, $Y$)\;
\KwRet{$R$}
}
\Fn{\FMain{$D_S, D_T, m, E$}}{
$m' \gets$ TRAIN($D_S, m, E$)\;
$R_T \gets$ EVALUATE($D_T, m'$)\;
\KwRet{$R_T$}
}
$R_T \gets$ TRANSFER\_KNOWLEDGE($D_S, D_T, m, E$)\;
\caption{Use Case 4 - Transfer learning evaluation}
\label{alg:transfer}
\end{algorithm}

\subsection{Use Case 5: Explainability Analysis}
\label{subsec:use-case-5}
\textit{Explainability} refers to the ability of a model to communicate the factors that influenced its decision-making process for a given input, particularly those that might lead the model to make incorrect decisions~\cite{adadi2018peeking, knyazev2019understanding}. Since deep-learning models are typically black-boxes, they are difficult to diagnose and adjust when failures occur. Thus, models which can better explain their decision-making process are easier to verify, debug, and make safer.
Our library enables users to analyze the explainability of different model architectures by visualizing the node attention scores of a graph learning model for a given input. 
The workflow of this use case is shown in Algorithm \ref{alg:explainability}. First, using a pre-trained graph learning model, we run inference on a dataset and record the model's spatial and temporal attention scores for each sequence to a CSV file. Then, we visualize the node attention scores for each \textit{scene-graph} and color code the nodes according to their attention score. For a given graph, the nodes with higher attention scores had a more significant impact on the decision made by the model. 

\begin{algorithm}[ht]
\SetAlgoLined
\LinesNumbered
\DontPrintSemicolon
\textbf{Input: A sequence of images from a driving video clip $I$, trained model $m$.}\;
\textbf{Output: Risk assessment result $\hat{Y}$, node attention scores $\alpha_t$ and temporal attention score $\beta_t$ for each graph in $G$.}\;
\SetKwFunction{Fsgvec}{SEQ2VEC\_ATTN}
\SetKwFunction{FMain}{GET\_ATTENTION\_SCORES}
\SetKwProg{Fn}{def}{:}{}
\Fn{\Fsgvec{$G$}}{
$h_G, \alpha \gets$ \{ \}, \{ \}\;
\For{$G_t$ in $G$}{
$\mathbf{h}_{G_t}, \alpha_{t} \gets$ \textsc{Spatial\_Model}($G_{t}$) \tcp*{$\alpha_{t}$ from SAGPool layer}} 
$Z, \beta \gets$ \textsc{Temporal\_Model}($\mathbf{h}_{G}$) \tcp*{$\beta$ from LSTM-attn layer}
$\hat{y}_0, \hat{y}_1 \gets$ \textsc{Activation}(MLP($Z$))\;
\uIf{$\hat{y}_1 \geq \hat{y}_0$}{ 
\KwRet $1, \alpha, \beta$\;}
\uElseIf{$\hat{y}_0 > \hat{y}_1$}{
\KwRet $0, \alpha, \beta$\;}
}
\Fn{\FMain{$I$}}{
$G \gets$ EXTRACT\_SEQ($I$)\;
$\hat{Y}, \alpha, \beta \gets$ SEQ2VEC\_ATTN($G$)\;
\KwRet{$\hat{Y}, \alpha, \beta$}\;
}
$\hat{Y}, \alpha, \beta \gets$ GET\_ATTENTION\_SCORES($I$)\;
\caption{Use Case 5 - Explainability analysis of \textit{scene-graph} risk assessment}
\label{alg:explainability}
\end{algorithm}
\section{Experiments}
In this section, we present results from running each use case presented in Section \ref{sec:usage} as well as details on the datasets and metrics used to evaluate each model.

\subsection{Dataset Preparation}
For experiments, we prepared two types of driving datasets: (i) synthesized lane-changing datasets (\textit{271-syn} and \textit{1043-syn}), and (ii) typical real-world driving datasets (\textit{571-honda} and \textit{1361-honda}). We labeled all of the datasets using our annotator tool as described in Section \ref{subsec:data.gen}. More details on the datasets as well as the labeling process can be found in \cite{yu2021scene}. 
We randomly split each dataset into a training set and a testing set by the ratio 7:3 such that the split is stratified, i.e., the proportion of risky to safe lane change clips in the training and testing sets is the same.
The models are first trained on the training set before being evaluated on the testing set.
The final score of a model on a dataset is computed by averaging over the testing set scores for five different stratified train-test splits.

\subsection{Model Configuration}
In our experiments, we use two graph learning architectures denoted MRGCN and MRGIN. Both models consist of the following structure: two graph convolution layers of size 64, one SAGPooling layer with 0.5 pooling ratio, one \textit{add} readout layer, and one problem-specific temporal model as defined in Figure \ref{fig:archi}. The two architectures only differ in the way successive graph convolutions are processed, as discussed in Section \ref{subsec:graph_learning}. As for the baselines, we evaluate the ResNet-50 CNN classifier and the CNN+LSTM classifier in our experiments. All models were evaluated using 5-fold cross-validation with the average test performance over the five folds presented as the final result.

\subsection{Use Case 1 Evaluation: \textit{Scene-Graph} Extraction}
\begin{figure}[ht]
    \centering
    \includegraphics[clip, trim=35 155 1 1, width=\textwidth]{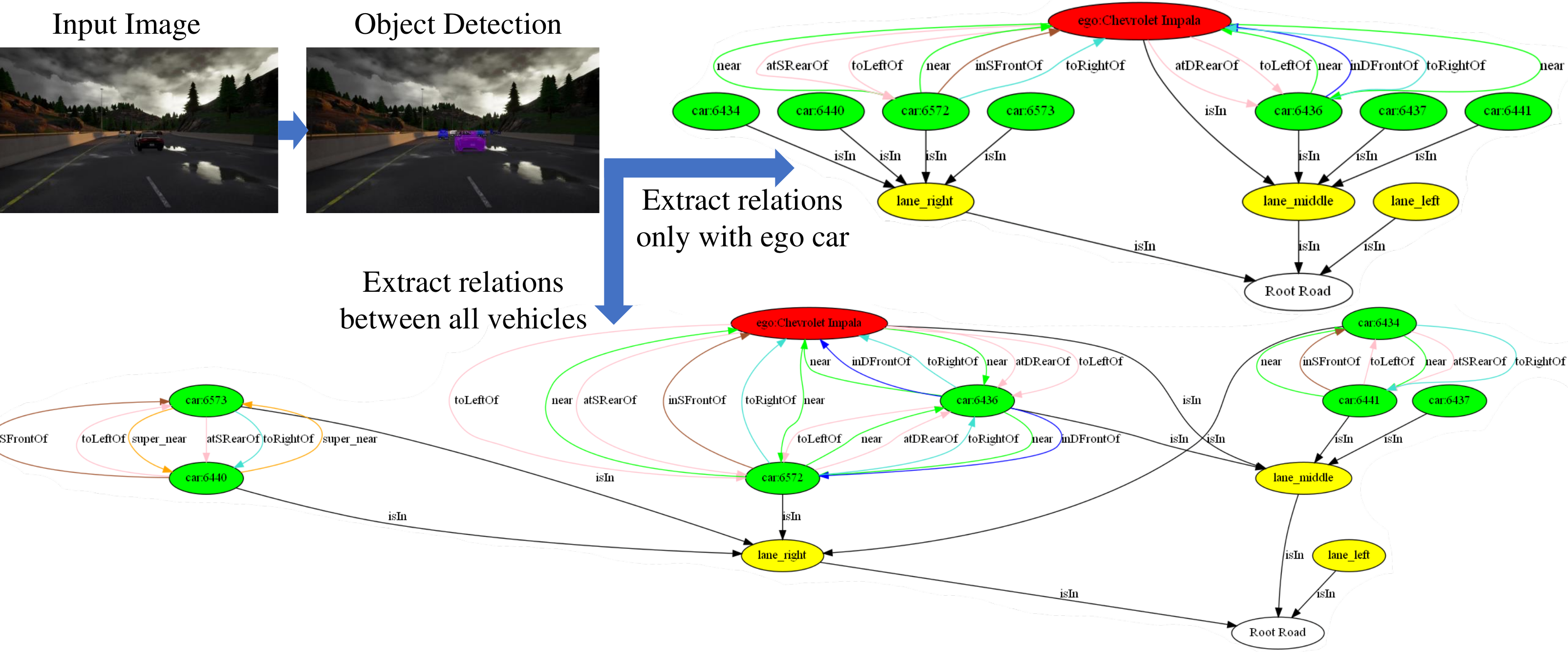}
    \caption{Demonstration of \textit{scene-graph} extraction with two different relation extraction settings. Zoom in for details.}
    \label{fig:usecase1}
\end{figure}

\begin{figure}
    \centering
    \includegraphics[width=.75\linewidth]{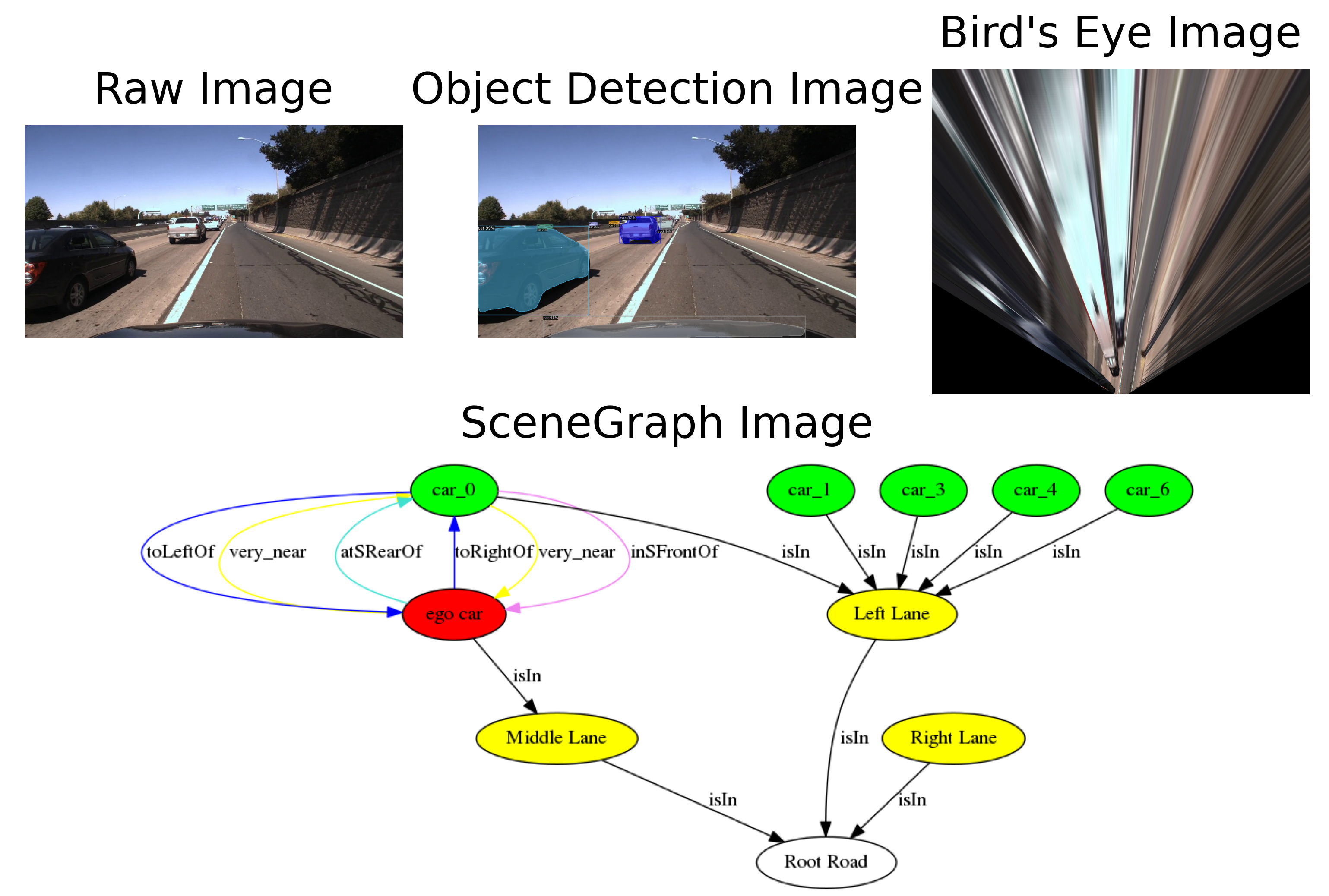}
    \caption{A demonstration of our \textit{scene-graph} visualization tool that enables the user to inspect: (i) an original input image, (ii) the object detection results, (iii) the birds-eye view projection of the image, and (iv) the resultant \textit{scene-graph}.}
    \label{fig:visualizer}
\end{figure}

In Figure \ref{fig:usecase1}, we show an example where two scene graphs are extracted from the same input image with different relation extraction settings. The graph at the bottom contains relations between all pairs of vehicles in the scene; for each pair of vehicles, if the two vehicles are within some distance threshold, the distance and direction relations are constructed. The graph at the top left is similar. However, it only contains relations between the ego vehicle and each other vehicle. This figure shows one example of how our tool enables flexible graph construction for different applications. A demonstration of our visualizer tool is shown in Figure \ref{fig:visualizer}. As shown, our visualizer allows the user to inspect how objects detected in the input image translate to the objects and relations in the \textit{scene-graph}.

\subsection{Use Case 2 Evaluation: Subjective Risk Assessment}
Here, we demonstrate how \textsc{roadscene2vec} can be used to train and evaluate several models for the subjective risk assessment use case.
We used classification accuracy and the Area Under the Curve (AUC) \cite{bradley1997use} of the Receiver Operating Characteristic (ROC) to score the models.
AUC, sometimes referred to as a balanced accuracy measure~\cite{sokolova2009systematic}, measures the probability that a binary classifier ranks a positive sample more highly than a random negative sample.
This is a more balanced measure for measuring accuracy, especially with imbalanced datasets (i.e., \textit{271-syn}, \textit{1043-syn}, \textit{571-honda}).

Table~\ref{tab:risk_result_1} shows a comparison between MRGCN, MRGIN, ResNet-50, and CNN+LSTM~\cite{yurtsever2019risky} models for driving scene risk assessment. The results show that the MRGCN based approach consistently outperforms the other models across all the datasets in terms of both classification accuracy and AUC. 
We found that the performance difference between the \textit{scene-graph} based approaches and the CNN-based approaches increased when the training datasets were smaller, indicating that the graph-based methods could likely learn a good representation with fewer data. 

\begin{table}[!ht]
    \centering
    \begin{tabular}{p{40pt} c c c c c c}
    \hline
        Metric & Dataset & MRGCN & MRGIN & ResNet-50 & CNN+LSTM~\cite{yurtsever2019risky}\\ \hline
    \multirow{4}{8pt}{Accuracy}& 271-syn & \textbf{0.9320} & 0.8561 & 0.6938 & 0.8033\\
&1043-syn & \textbf{0.9580} & 0.8784 & 0.9053 & 0.7742\\
&571-honda & \textbf{0.8710} & 0.8310 & 0.7689 & 0.6041\\
&1361-honda & \textbf{0.8655} & 0.7245 & 0.6839 & 0.7158\\
    \hline
    \multirow{4}{8pt}{AUC} & 271-syn & \textbf{0.9620} & 0.9437 & 0.7371 & 0.8394\\
&1043-syn & \textbf{0.9780} & 0.9591 & 0.9616 & 0.8221\\
&571-honda & \textbf{0.9105} & 0.8903 & 0.8343 & 0.6670\\
&1361-honda & \textbf{0.9124} & 0.8164 & 0.7340 & 0.7560\\
        \hline
    \end{tabular}
    \caption{Risk assessment result for MRGCN, MRGIN, ResNet-50, and CNN+LSTM.}
    \label{tab:risk_result_1}
\end{table}

\subsection{Use Case 3 Evaluation: Collision Prediction}
Next, we evaluated the models in \textsc{roadscene2vec} at collision prediction using classification accuracy, AUC, and Matthews Correlation Coefficient (MCC) \cite{chicco2020advantages}. 
MCC is considered a balanced performance measure for binary classification, even on datasets with significant class imbalances. The MCC score outputs a value between -1.0 and 1.0, where 1.0 corresponds to a perfect classifier, 0.0 to a random classifier, and -1.0 to an always incorrect classifier. The results from our evaluation are shown in Table \ref{tab:usecase3results}. 

Once again, MRGCN outperforms the other models on the synthetic datasets. However, on the 571-honda dataset, the ResNet-50 model outperforms MRGCN across all metrics. Upon deeper inspection of the results, we found that the ResNet-50 model had a higher FNR than the MRGCN and a lower FPR than the MRGCN, suggesting that the ResNet-50 model is less sensitive than the MRGCN. Given that collision prediction is a safety-critical application, this behavior may not necessarily be desirable; however, decision boundary tuning could be used to fine-tune the sensitivity for the final application's requirements. 

On both Use Case 2 and 3, MRGIN underperforms MRGCN, likely because MRGCN is a more general framework while MRGIN is designed to perform well at graph topology analysis problems, such as graph isomorphism testing. MRGIN may outperform MRGCN on different problem formulations or graph construction formulations if they play to these strengths of MRGIN. 

\begin{table}
    \centering
    \begin{tabular}{p{40pt} c c c c c c}
    \hline
    Metric & Dataset & MRGCN & MRGIN & ResNet-50 & CNN+LSTM \cite{yurtsever2019risky}\\\hline
    \multirow{3}{8pt}{Accuracy} 
&271-syn & \textbf{0.8812} & 0.8028 & 0.7039 & 0.7184\\
&1043-syn & \textbf{0.9095} & 0.7803 & 0.8080 & 0.8029\\
&571-honda & 0.6922 & 0.7230 & \textbf{0.7340} & 0.5606\\
     \hline
    \multirow{3}{8pt}{AUC} 
&271-syn & \textbf{0.9457} & 0.8724 & 0.7564 & 0.7607\\
&1043-syn & \textbf{0.9477} & 0.8826 & 0.9026 & 0.8493\\
&571-honda & 0.7775 & \textbf{0.7844} & 0.7802 & 0.5871\\
     \hline
    \multirow{3}{8pt}{MCC} 
&271-syn & \textbf{0.5145} & 0.3046 & 0.3320 & 0.1474\\
&1043-syn & \textbf{0.5385} & 0.2852 & 0.4602 & 0.2436\\
&571-honda & 0.2142 & 0.1908 & \textbf{0.3547} & 0.1347\\
     \hline
    \end{tabular}
    \caption{Collision prediction accuracy, AUC, and MCC for different models in \textsc{roadscene2vec}.}
    \label{tab:usecase3results}
\end{table}

\subsection{Use Case 4 Evaluation: Transfer Learning}
Here, we demonstrate how \textsc{roadscene2vec} can be used to evaluate each model's ability to transfer the knowledge learned from simulated datasets to real-world datasets.
As part of this use case, \textsc{roadscene2vec} uses the model weights and parameters learned from training on the simulated dataset (\textit{271-syn} or \textit{1043-syn} in this case) directly for testing on the real-world driving dataset (\textit{571-honda}) with no domain adaptation steps. 
We show the results of this evaluation for the MRGCN, ResNet-50, and CNN+LSTM models in Table~\ref{tab:transfer}. 

As expected, the performance of all models degrades when tested on \textit{571-honda} dataset. 
However, as Table~\ref{tab:transfer} shows, the accuracy of the MRGCN only drops by 3.5\% and 6.5\% when the model is trained on \textit{271-syn} and \textit{1043-syn}, respectively, while the CNN+LSTM's performance drops by 27.9\% and 17.3\%, respectively. Furthermore, the MRGCN achieves a higher accuracy score than the CNN+LSTM when transferring from the smaller \textit{271-syn} dataset, once again indicating that \textit{scene-graph} models can better model the problem even when trained on smaller amounts of data.
The ResNet-50 model performs worst and classifies most of the sequences as risky, resulting in an accuracy score nearly equivalent to the proportion of risky sequences in the \textit{571-honda} dataset (~17.25\%).
These results suggest that the \textit{scene-graph} models can transfer knowledge more effectively than the CNN-based models.

\begin{table}[ht]
\centering
\begin{tabular}{p{95pt} p{75pt} p{47pt} p{80pt}}
\hline
Experiment & Model & Original Acc. & Transfer Acc. \\\hline
\multirow{3}{*}{\textit{271-syn}~to \textit{571-honda}} 
&ResNet-50 & 0.7039 & 0.1899 (-0.514)\\
& CNN+LSTM~\cite{yurtsever2019risky} & 0.8033 & 0.5244 (-0.279)\\
& \textbf{MRGCN} & \textbf{0.9040} & \textbf{0.8690 (-0.035)}\\
\hline
\multirow{3}{*}{\textit{1043-syn}~to \textit{571-honda}}
& ResNet-50 & 0.8080 & 0.1725 (-0.636)\\
& CNN+LSTM~\cite{yurtsever2019risky} & 0.7742 & 0.6010 (-0.173)\\
& \textbf{MRGCN} &\textbf{0.9520} &\textbf{0.8870 (-0.065)}\\
\hline
\end{tabular}
\caption{The results of comparing transferability between MRGCN, ResNet-50, and CNN+LSTM  \cite{yurtsever2019risky}. In this experiment, we trained each model on both the \textit{271-syn} dataset and \textit{1043-syn} dataset.
Then we evaluated the accuracy of the trained model on both original dataset and \textit{571-honda} dataset without any domain adaptation.}
\label{tab:transfer}
\end{table}

\subsection{Use Case 5 Evaluation: Explainability Analysis}
\begin{figure}
    \centering
    \includegraphics[clip, trim=0 245 0 10, width=\linewidth]{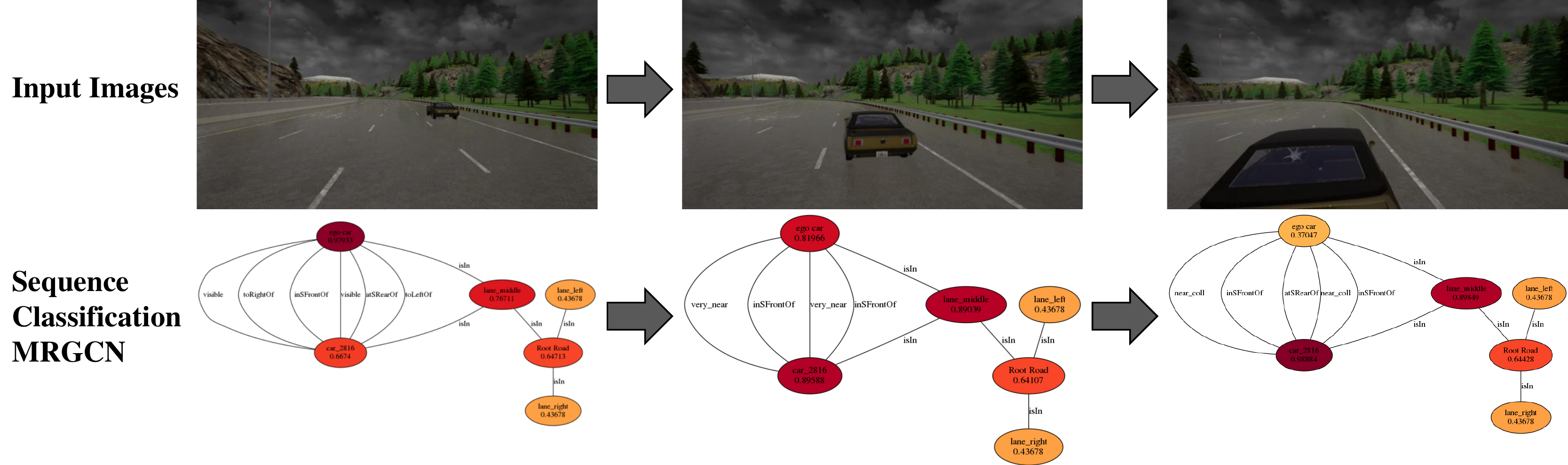}
    \caption{A demonstration of how Use Case 5 enables \textit{explainability} analysis. For this driving sequence, it can be clearly seen how the node attention scores shift to give higher weight to the approaching vehicle as its distance to the ego car reduces.}
    \label{fig:usecase5}
\end{figure}

To demonstrate \textsc{roadscene2vec}'s tools for evaluating explainability, we run our explainability analysis tool on the MRGCN model trained for risk assessment on the 271-syn dataset. The result from analyzing one of the sequences from the dataset is shown in Figure \ref{fig:usecase5}. As shown, the attention scores are highest on the nodes which present the highest degree of risk. Additionally, the graph with the highest attention score for the other vehicle is also the graph corresponding to the collision with the other vehicle.

\section{Discussion}
\label{sec:discussion}

\subsection{Practicality}
Although \textsc{roadscene2vec} is intended to be a tool that benefits the research community, its practicality and carryover to real-world applications are equally important. As shown with Use Case 4, \textsc{roadscene2vec} enables researchers to directly evaluate the ability of models trained on synthetic data to transfer their knowledge to real-world driving scenes. Many research papers often overlook this critical problem, leading to a disconnect between simulated trials and real-world performance. Our tool better enables the study of this crucial problem area and allows researchers to analyze the real-world practicality of various graph-based methodologies. Furthermore, we show that \textsc{roadscene2vec} is directly compatible with both the real-world Honda driving dataset \cite{Ramanishka_behavior_CVPR_2018} as well as the popular open-source driving simulator, CARLA \cite{dosovitskiy2017carla}, making our tool useful for a wide range of AV applications.

\subsection{Limitations and Future Work}
Although \textsc{roadscene2vec} provides a suite of tools for training and evaluating both scene-graph-based and CNN-based models, there are some limitations to its capabilities. For example, \textsc{roadscene2vec} currently only supports input data in the format of ground-truth data from the CARLA simulator or image data from a forward-facing camera; it currently does not support radar, lidar, or multi-camera data. We selected image data and CARLA data as the primary input modalities because these data types are the ones most used by AV researchers currently. Although radar and lidar data are valuable and well-studied in specific applications such as localization and sensor fusion, most AV research papers exploring perception and control methodologies use camera-based inputs. However, this limitation can be overcome by implementing preprocessors for extracting (or fusing) \textit{scene-graphs} from these different modalities. Thus, \textsc{roadscene2vec} does not currently support multiple sensing modalities but could support them as part of future work. 
Furthermore, our tool does not implement more than a few common perception algorithms and use cases. However, our tool is designed to be modular and re-configurable to support custom models and problem formulations. We expect that researchers will develop custom architectures and models for the various well-studied problems in the AV domain and provide instructions in our repository for integrating the custom models with \textsc{roadscene2vec}'s workflow. Thus, we leave the study of other AV applications and model architectures as future work. \green{ We also welcome outside contributions to our open-source tool to further improve its utility for the research community.}

\section{Conclusion}
\label{sec:conclusion}
It is clear from current research as well as the examples shown in this paper that \textit{scene-graph} representations of road scenes can be beneficial for a wide range of AV applications. In this paper, we introduced and demonstrated our tool for exploring and studying the applications of road \textit{scene-graphs}, named \textsc{roadscene2vec}. We showed that our re-configurable graph-construction methodology enables the study of different graph layouts for various problems. We also demonstrated performance evaluations for conventional CNN architectures and graph-based models for two common AV perception use cases: risk assessment and collision prediction. Furthermore, we showed how our tool facilitates studying the transferability and explainability of graph-based AV models for both synthetic and real-world data. We believe our open-source tool fills a significant gap in the research community and will enable a deeper study of the applicability and practicality of graph-based solutions for AV problems.


\section*{Acknowledgment}
This work was partially supported by the National Science Foundation (NSF) under award CMMI-1739503 and by Graduate Assistance in Areas of National Need (GAANN) under award P200A180052. 
Any opinions, findings, conclusions, or recommendations expressed in this paper are those of the authors and do not necessarily reflect the views of the funding agency.

\bibliographystyle{ACM-Reference-Format}
\bibliography{bibliography.bib}
\end{document}